%% file: main.tex
\documentclass[preprint,12pt]{elsarticle}
\usepackage{amsmath,amssymb,booktabs,tabularx,graphicx}
\usepackage[hidelinks]{hyperref}
\usepackage{verbatim}

\biboptions{numbers,sort&compress}
\newcommand{\nMAE}{\operatorname{nMAE}}

\journal{Energy}

\begin{document}
\begin{frontmatter}

\title{State transport routing for short-horizon adaptation in multi-horizon photovoltaic forecasting}

\author[1]{Xu Yuqing}
\author[1]{Zhou Liguo}
\author[1]{Sun Ze}
\author[1]{Yu Lei}
\author[1]{Jiang Mingming}
\affiliation[1]{organization={Huaibei Normal University},country={China}}

\begin{abstract}
Recent power measurements provide valuable information for photovoltaic (PV) power forecasting, but directly extrapolating short-term trends can introduce substantial errors over longer forecast horizons. To address this challenge, we propose state transport routing (STR), a lightweight adapter that refines the predictions of a frozen forecasting model. STR combines the original forecast with two complementary trajectories derived from the latest measured power level and its recent trend. A horizon-conditioned router adjusts their contributions over the first 120 min, while leaving subsequent predictions unchanged. Experiments on four public PV datasets show that STR consistently outperforms a parameter-matched residual adapter. On PVDAQ, the same approach improves five neural forecasting backbones, reducing all-horizon normalized mean absolute error by 0.0201--0.2364 percentage points, with paired 95\% confidence intervals excluding zero. No reliable improvement is observed for LightGBM. These findings demonstrate the potential of structured state adaptation to improve short-term forecasting across different neural architectures without retraining the underlying models or altering their longer-horizon predictions.
\end{abstract}

\begin{keyword}
Photovoltaic power forecasting \sep
Multi-horizon forecasting \sep
State transport routing \sep
Forecast adaptation \sep
Forecast post-processing
\end{keyword}
\end{frontmatter}

\section{Introduction}
Accurate photovoltaic (PV) power forecasting is essential for power system operation and energy management. For example, day-ahead work has combined on-site weather observations with numerical forecasts \cite{luo2023multifidelity}, while ultra-short-term work has used historical power and meteorological data \cite{huang2023mlstnet}. Recent power measurements provide valuable information about the current operating state, allowing forecasting models to respond to rapid changes in power generation. However, short-term fluctuations may not persist over longer horizons, and directly extrapolating recent trends can lead to substantial prediction errors \cite{antonanzas,inman}. Therefore, multi-horizon forecasting models need to incorporate recent operating information while preserving the reliability of their longer-term predictions.

Recent forecasting models capture temporal dynamics in different ways.
N-HiTS and TimeMixer use hierarchical interpolation and multiscale
mixing, respectively, while PatchTST and iTransformer employ
attention-based representations
\cite{nhits,timemixer,patchtst,itransformer}.
Simpler architectures, such as DLinear and TSMixer, offer alternative
trade-offs between forecasting accuracy and computational cost
\cite{dlinear,tsmixer}.
Rather than modifying these forecasting architectures, this study
focuses on adapting the output of an already trained model:
recent power measurements are used to refine its short-horizon
predictions, while longer-horizon predictions remain unchanged.

Simple state-based corrections provide useful short-term information, but each has clear limitations. Persistence keeps the forecast close to the latest observation, while a local-slope extrapolation extends the recent trend into the future. The former ignores subsequent evolution, whereas the latter may overextend a transient ramp. A generic residual adapter offers greater flexibility, but does not explicitly distinguish between these different forecasting assumptions. This motivates an adaptation mechanism that can select among them at each forecast horizon while limiting its influence to the short term.

To address this problem, we propose state transport routing (STR),
which adapts the predictions of a pretrained forecasting model
without modifying its parameters. STR considers three candidate
trajectories: the original model prediction, the latest observed
power level, and an extrapolation of the recent power trend.
A horizon-conditioned router combines these trajectories and
learns an additional residual correction for the first 120 min.
Beyond this interval, the predictions remain identical to those
of the original model. STR operates directly on the forecast
output using causally available information, without requiring
access to the backbone's internal representations.

We evaluate STR by addressing three research questions:
(RQ1) Does explicit state routing provide an advantage over
a conventional residual correction?
(RQ2) Can the same STR design improve different frozen neural
forecasting models?
(RQ3) How does its effectiveness vary across forecast horizons,
model architectures, and operating conditions?

The main contributions of this study are as follows:
\begin{enumerate}
    \item We propose STR, a lightweight adapter that combines
    the original model forecast with two trajectories derived
    from recent power measurements. Horizon-conditioned routing
    adjusts their contributions over the first 120 min, while
    leaving subsequent predictions unchanged.

    \item We evaluate STR on four public PV datasets and compare
    it with a nominally parameter-matched residual adapter.
    The results show that explicitly incorporating state
    trajectories improves forecasting accuracy beyond the
    conventional residual correction used in this comparison.

    \item We investigate the applicability of STR to five
    different neural forecasting models under a common PVDAQ
    protocol. We also examine its limitations, including the
    lack of a reliable improvement for LightGBM and the
    performance degradation observed under low-volatility
    conditions in Solar-Energy.
\end{enumerate}

\section{Related work}
\subsection{Multi-horizon PV forecasting}

Multi-horizon PV forecasting requires models to capture both
the daily variation in solar generation and short-term changes
caused by weather and operating conditions. Solar geometry
provides information about the daily generation pattern
\cite{spa,pvlib}, while weather forecasts can provide additional
information about future conditions.

To model these dynamics, previous PV forecasting studies have
explored Transformer-based and recurrent architectures
\cite{piantadosi2024,tao2024,kim2024,ma2024}.
Other forecasting models, including N-HiTS, TimesNet, TimeMixer,
TiDE and TimeXer, employ hierarchical, multiscale or
exogenous-variable modeling strategies
\cite{nhits,timesnet,timemixer,tide,timexer}.
Recent studies in \emph{Energy} have also investigated stacking
and temporal-scale decoupling to improve forecasting performance
\cite{cao,spinet2026}.

While these studies focus on improving the forecasting models
themselves, the present work considers how their predictions
can be adapted after training. Rather than developing another
forecasting architecture, STR uses recent power measurements
to refine short-horizon predictions while preserving the
original forecasts at longer horizons.

\subsection{Forecast combination and short-range state information}

Recent power measurements provide a simple basis for short-term
forecasting. Persistence assumes that the latest observed power
level remains unchanged, while trend extrapolation extends recent
power variations into the future \cite{antonanzas,inman}.
Although these approaches can capture the current operating state,
their assumptions may become less reliable as the forecast horizon
increases.

Forecast combination and stacking offer another way to improve
prediction accuracy by exploiting complementary forecasts
\cite{bates,wolpert,breiman}. Similar strategies have been explored
in PV forecasting through combinations of neural and tree-based
models \cite{zhang2022hybrid,xiong2024gsk}.

Rather than combining multiple independently trained forecasting
models, STR combines the output of a single frozen backbone with
two simple trajectories derived from recent power measurements.
Their contributions are adjusted according to the forecast
horizon, and the resulting correction is restricted to the
short-term prediction interval.

\subsection{Output adaptation and positioning of STR}

PV forecasting has used error-correction stages alongside a primary
predictor \cite{zhang2022hybrid}. Such a stage can refine an output
without changing the primary model. Stacking, by contrast, combines
several complete forecasts \cite{cao,bates,wolpert}. Our comparison
uses a single frozen predictor and an additive residual adapter as
its control; that control does not expose distinct assumptions about
the latest level or trend.

STR keeps this output interface and adds two state trajectories to
the frozen forecast. A horizon-conditioned router combines the three
paths and adds a residual correction through 120 min; later outputs
return the backbone prediction exactly.

To examine the contribution of the explicit state trajectories,
we compare STR with a nominally parameter-matched residual adapter
under the same experimental protocol.

\section{Method}

\subsection{Problem formulation and design objective}

At each forecast origin $t$, the available power measurements are
denoted by $\mathbf{x}_t=(x_{t-L+1},\ldots,x_t)$, where $L$ is the
historical input length. Let $\mathbf{c}_t$ represent the covariates
available at the time of prediction. A pretrained forecasting model,
referred to as the backbone, generates an $H$-step forecast:
\begin{equation}
    \mathbf{b}_t=f_{\theta}(\mathbf{x}_t,\mathbf{c}_t).
\end{equation}

The backbone parameters $\theta$ are kept fixed throughout adaptation.
STR operates on its predictions rather than modifying the forecasting
model itself. Its objective is to incorporate recent power information
into short-horizon predictions while preserving the original forecast
at longer horizons.

\subsection{Candidate state transports}

STR constructs three candidate trajectories: the frozen
backbone forecast and two state-derived paths that carry
the latest observed power forward, either unchanged or
according to its recent trend. For forecast step $h$,
they are defined as
\begin{align}
    p^{(1)}_{t,h}&=b_{t,h},\\
    p^{(2)}_{t,h}&=x_t,\\
    p^{(3)}_{t,h}&=x_t+h\frac{x_t-x_{t-3}}{3}.
\end{align}

The first trajectory retains the backbone prediction, which describes
the future evolution learned from historical data and available
covariates. The second assumes that power remains at its latest
observed level, whereas the third extends the recent power trend
into the future. Together, these trajectories provide complementary
alternatives for adapting the original forecast.

The slope-based trajectory is a simple extrapolation of recent
observations rather than a physical model of power ramps. Here,
$h$ denotes the number of forecast steps, so the extrapolation
follows the native sampling interval of each dataset.

To determine how these trajectories should be combined, STR uses
a local-state descriptor $\mathbf{u}_t$ containing the latest 12
power measurements, recent differences and local variability.
Station identity is also included for pooled datasets. The forecast
horizon is represented separately by a learned embedding.

The available information differs across datasets. For GEFCom,
the descriptor additionally includes the first two forecast-predictor
vectors and their change. On PVDAQ, solar geometry and archived GFS
forecasts are supplied to the frozen backbone, whereas the core
router uses recent power information and the horizon embedding.
Gatton and Solar-Energy use power history without additional
weather inputs. Thus, direct weather conditioning is not required
for the core STR formulation.

\subsection{Horizon-conditioned routing and exact bypass}

The local-state descriptor is first projected into a hidden
representation and combined with the horizon embedding:
\begin{equation}
    \mathbf{z}_{t,h}
    =\operatorname{GELU}(W_u\mathbf{u}_t+\mathbf{e}_h).
\end{equation}

A linear output layer then produces three routing logits
$\mathbf{a}_{t,h}\in\mathbb{R}^{3}$ and an additive residual
$d_{t,h}$. The routing weights are obtained using a softmax
function,
$\boldsymbol{\alpha}_{t,h}
=\operatorname{softmax}(\mathbf{a}_{t,h})$.

For horizons within the adaptation window, the final prediction
is given by
\begin{equation}
    \hat y_{t,h}
    =\sum_{j=1}^{3}
    \alpha^{(j)}_{t,h}p^{(j)}_{t,h}+d_{t,h},
    \qquad h\leq K.
    \label{eq:str}
\end{equation}

The routing weights determine the contribution of each candidate
trajectory at every forecast step, while the residual term provides
an additional learned correction.

Adaptation is restricted to the first 120 min. For all subsequent
horizons, STR directly returns the original backbone prediction:
\begin{equation}
    \hat y_{t,h}=b_{t,h},
    \qquad h>K.
    \label{eq:bypass}
\end{equation}

\begin{figure}[!t]
    \centering
    \includegraphics[width=\linewidth]{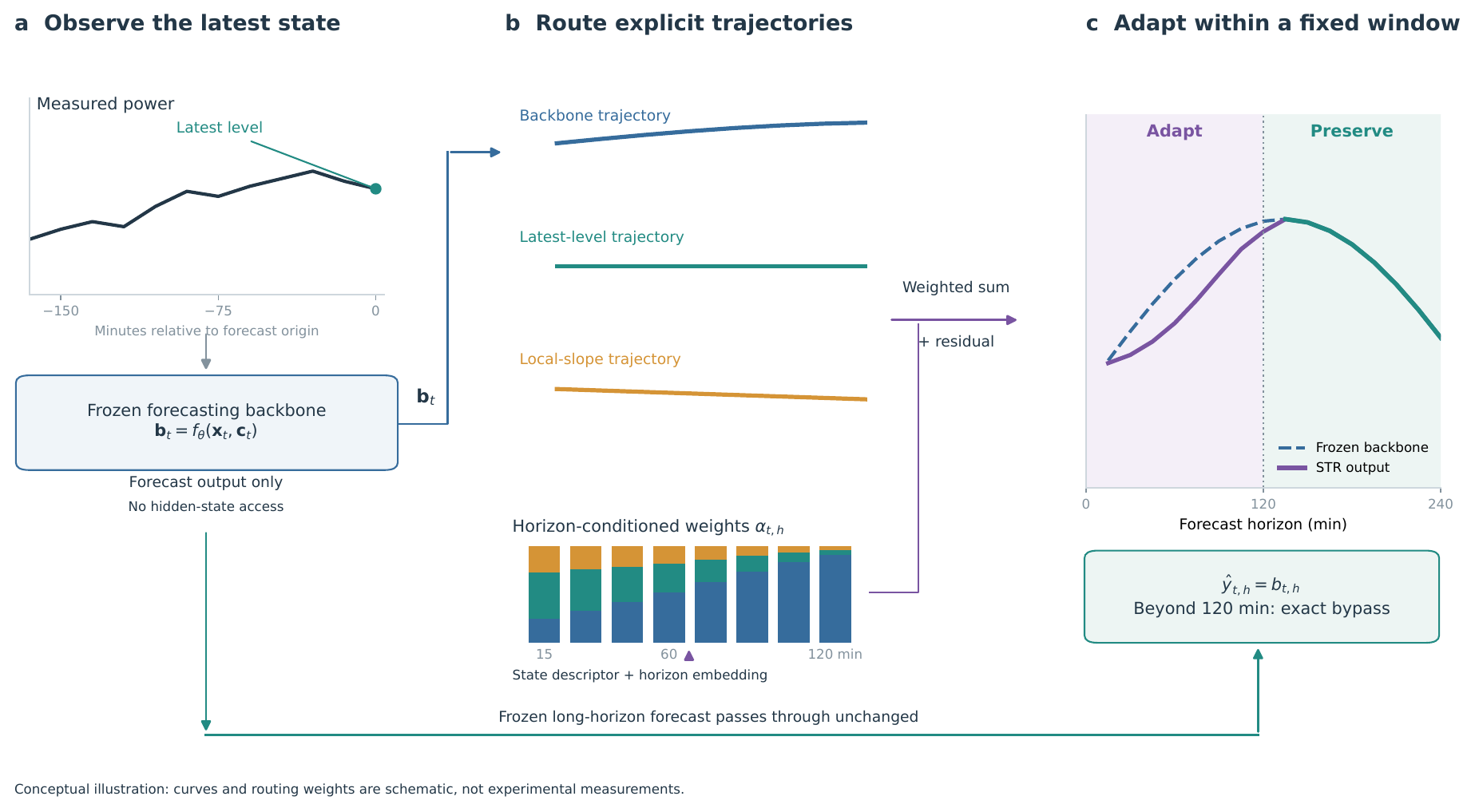}
    \caption{Overview of STR. (a) The backbone forecast and recent
    power observations are used to construct three candidate
    trajectories. (b) A horizon-conditioned router combines these
    trajectories and adds a residual correction. (c) Predictions
    are adapted within the first 120 min, while subsequent
    backbone outputs are preserved exactly. The illustrated
    trajectories and routing weights are schematic.}
    \label{fig:architecture}
\end{figure}

Here, $K$ is the forecast step corresponding to 120 min at the
native sampling interval. This design allows STR to adjust
short-horizon predictions without changing the backbone output
for the third and fourth hours. The complete architecture is
illustrated in Fig.~\ref{fig:architecture}.

\subsection{Matched mechanistic control}

To examine whether the explicit state trajectories contribute
to the improvement, we compare STR with a residual adapter
using the same descriptor, hidden projection, nominal parameter
count and optimization settings. The control removes the
candidate-state combination and instead learns an additive
correction to the backbone forecast:
\begin{equation}
    \hat y^{\mathrm{control}}_{t,h}
    =b_{t,h}+d_{t,h},
    \qquad h\leq K.
\end{equation}

Both methods use the same long-horizon bypass defined in
Eq.~\eqref{eq:bypass}. This comparison isolates the contribution
of the explicit trajectories within the implemented adapter
architecture. Because the routing logits are inactive in the
control, matching the nominal parameter count does not imply
identical effective capacity for every possible residual-adapter
design.

\subsection{Backbones and optimization}

STR is trained on top of a pretrained forecasting backbone,
whose parameters remain fixed during adaptation. For GEFCom,
we use a compact forecast-conditioned convolutional network
with dilation rates of $1$, $2$, $4$ and $8$. Gatton,
Solar-Energy and PVDAQ use TimeMixer as the backbone.
The corresponding STR adapters contain 2,312, 1,628,
2,780 and 1,204 trainable parameters, respectively.
For the three TimeMixer-based models, these account for
only 0.047\%, 0.077\% and 0.034\% of the backbone
parameter counts.

All adapters are trained for 24 epochs using AdamW,
with a learning rate of $10^{-3}$, weight decay of
$10^{-4}$ and a batch size of 256. Gradient clipping
is applied with a maximum norm of one. We evaluate
the models on the selection set every two epochs
and retain the checkpoint with the lowest MAE.

To account for training variability, we use three random
seeds (2021--2023). GEFCom, Gatton and Solar-Energy
include three backbone seeds, whereas PVDAQ uses a single
pretrained backbone checkpoint with three independently
trained adapters. Therefore, the variation across PVDAQ
runs reflects adapter training rather than variability
from retraining the entire forecasting model.

\subsection{Backbone interface and transfer protocol}

STR operates on the forecast output of a pretrained model,
using only its prediction vector $\mathbf{b}_t\in\mathbb{R}^{H}$
and causally available state information. It does not require
access to the backbone's hidden representations, and the
backbone parameters remain frozen during adaptation.

To examine its applicability across different forecasting
architectures, we apply the same three-trajectory design and
routing mechanism to each backbone. A separate STR adapter
is trained for each model, without sharing learned router
weights between architectures. This experiment therefore
evaluates whether the STR design can be reused across
different models, rather than whether a trained adapter
can be transferred directly between them.

The common PVDAQ evaluation protocol is described in
the following section.

\section{Experimental design}
\subsection{Datasets and chronological partitions}
Table~\ref{tab:data} defines the four tasks. Native targets and normalizations are retained because capacity metadata and sampling structure differ across datasets.

\begin{table}[!htbp]
\centering\small
\setlength{\tabcolsep}{2.2pt}
\caption{Public datasets and frozen forecasting tasks. Counts are retained training/selection/test forecast-origin windows, one per series and origin.}
\label{tab:data}
\begin{tabularx}{\linewidth}{@{}l@{\hspace{8pt}}rrrr>{\raggedright\arraybackslash}X@{}}
\toprule
Dataset & Series & Step & Horizons & Origins & Available information \\
\midrule
GEFCom & 3 & 60 min & 4 & 34,458/5,793/1,887 & Power, calendar, 12 issued predictors \\
Gatton & 1 & 15 min & 16 & 16,897/2,961/5,613 & Power history; 3.275 MW nameplate \\
Solar-Energy & 137 & 10 min & 24 & 32,743/39,593/239,613 & Per-series power history \\
PVDAQ 2107 & 1 & 15 min & 16 & 139,869/17,197/29,211 & Power, solar geometry, archived GFS \\
\bottomrule
\end{tabularx}
\end{table}

The four datasets are divided chronologically into training,
selection and test sets. For GEFCom, the training period
covers April 2012 to September 2013, followed by a selection
period from October to December 2013 and a test period in
June 2014. Gatton uses January--June 2020 for training,
July 2020 for selection and September--October 2020 for
testing. Solar-Energy uses the first 70\% of its 52,560
ten-minute source rows for training and the final 20\%
for testing. The intervening 10\% is divided into three
equal chronological blocks; Table~\ref{tab:data} reports
the first as selection, while the other two are reserved
for the source protocol's inner and outer fusion stages.
For PVDAQ 2107, the corresponding periods are 2019--2022,
2023 and January--October 2024.

Solar-Energy's table counts are eligible (series, forecast-origin)
windows, not raw time rows. Each window has 144 history steps
and 24 target steps, with origins spaced six rows (one hour) apart.
The training block contains 836,385 eligible windows; the frozen
stratified cap of 239 per series retains 32,743 for fitting.
Selection and test retain all 39,593 and 239,613 eligible windows,
respectively. The window counts therefore do not follow the
raw-row split percentages.

Dataset-specific preprocessing and evaluation conventions
are retained. Gatton's one-minute measurements are aggregated
into 15-minute means using complete intervals, following a
sign audit performed on the training data only. Because
verified plant capacities are unavailable for Solar-Energy,
performance is evaluated using train-standardized macro MAE.
For PVDAQ, power is normalized by the reported 893 kW DC
nameplate capacity, and the daylight metric is restricted
to solar elevations above $5^{\circ}$.

\subsection{Mechanism evaluation}

To assess the contribution of the three candidate trajectories,
we compare STR with a residual adapter on all four datasets.
Both methods use the same frozen backbone predictions,
state descriptor, training procedure and long-horizon bypass.
The residual adapter removes the trajectory combination and
instead learns an additive correction to the original forecast.
Their nominal parameter counts are also matched.

This comparison examines whether explicit state routing provides
an advantage over the residual correction used in this study,
independently of STR's performance relative to other forecasting
models.

\subsection{Backbone-transfer evaluation}

To investigate whether STR can be applied to different forecasting
architectures, we evaluate it on five neural backbones using the
same PVDAQ dataset and experimental protocol. These include
TimeMixer, N-HiTS, TSMixer, DLinear and iTransformer. All models
use the same power history, solar geometry, historical GFS
information and archived future GFS forecasts. LightGBM is
included to examine whether the adaptation approach also
benefits a tree-based forecasting model. Its input features
are derived from the same raw information. All backbone
parameters and predictions are kept fixed throughout the
experiment.

For each backbone, we independently train an STR adapter
and a nominally parameter-matched residual adapter.
Both receive the frozen 16-step forecast and the latest
12 power observations. The STR architecture, routing
window ($K=8$), optimizer, learning rate, batch size
and 24-epoch training schedule are kept unchanged
across models. We use the same three random seeds
(2021--2023), with a separate adapter trained for each
backbone and seed. The chronological data split,
29,211 test origins and evaluation masks are also
identical across experiments. All adapters are frozen
before their test predictions are generated, and no
backbone-specific tuning is performed to improve
the transfer results.

The backbone scores in this experiment differ slightly
from those reported in the external benchmark because
the two comparisons use different seed-aggregation
procedures. Here, TSMixer, iTransformer and DLinear
use their original selected seed-2021 checkpoints,
with baseline nMAE values of 4.9492\%, 4.9934\%
and 5.7695\%, respectively. In contrast, the external
benchmark reports the mean performance of three
independently trained backbone seeds, giving
4.9427\%, 5.0252\% and 5.7550\%. Both sets of
results are retained, as they refer to the same
data preprocessing and forecast origins but different
aggregation procedures.

\subsection{External benchmark}

To assess the forecasting performance of STR against existing
methods, we conduct a common benchmark across four public PV
datasets. The comparison includes 13 learning-based methods:
LightGBM, XGBoost, CatBoost, N-HiTS, TimeMixer, STR, PatchTST,
iTransformer, TimeXer, TiDE, TimesNet, DLinear and TSMixer.
Within each dataset, all methods follow the same chronological
data split and use identical forecast origins, history lengths,
prediction targets and available input information.

Seven additional neural architectures are implemented using
the official Time-Series-Library repository at commit
\texttt{4e938a1}, while XGBoost and CatBoost use their released
Python packages. Learning rates and model checkpoints are
selected exclusively on the selection set. All newly trained
methods are evaluated with three random seeds (2021--2023),
and all reported results are obtained from executed models
rather than copied from published studies.

AMPDNet and PV-Client are included in the GEFCom comparison,
where their available implementations support the corresponding
data interface. Dataset-specific naive forecasts are also
retained as reference methods. The frozen backbone and residual
adapter are used only for the mechanism evaluation and are not
included in the external benchmark rankings.

The GEFCom H0--H3 weather experiment is reported separately,
as it involves retraining six methods with a common expanded
weather-input interface. Inference latency is reported separately
for CPU-based tree models and GPU-based neural models to avoid
direct comparisons across different hardware configurations.

\subsection{Metrics and paired uncertainty}

For datasets with capacity-normalized targets, forecasting error
is measured using normalized mean absolute error (nMAE):
\begin{equation}
    \nMAE_h =
    \frac{1}{N_h}
    \sum_{t=1}^{N_h}
    \frac{|P_{t,h}-\hat P_{t,h}|}{C}.
\end{equation}
Here, $P_{t,h}$ and $\hat P_{t,h}$ denote the observed and
predicted power at forecast horizon $h$, respectively.
$N_h$ is the number of evaluated predictions, and $C$
is the corresponding normalization capacity.
Gatton uses a 3,275 kW nameplate denominator for its net-export
proxy, and PVDAQ uses the reported 893 kW DC nameplate for measured
AC output. GEFCom supplies already normalized power, so its error
is MAE on the published unit-scale target; we do not infer an
unreported physical capacity for its three zones.

For Solar-Energy, each of the 137 series is centered and divided by
its own training-period standard deviation. We calculate MAE within
each series at each horizon and then give the series equal weight.
Its train-standardized macro MAE is therefore not capacity-normalized
nMAE. Let $E_{d,h}$ denote the resulting dataset-specific scaled MAE:
published normalized-target MAE for GEFCom, capacity-normalized MAE
for Gatton and PVDAQ, and train-standardized macro MAE for Solar-Energy.
The reported all-horizon error is the unweighted mean of $E_{d,h}$.
The tabulated score $100(1-E_d)$ is a descriptive transformation;
Solar-Energy's 86.517\% is not an engineering accuracy relative to
plant capacity. Errors are compared within each dataset, not as
interchangeable physical percentages across datasets.

We use paired temporal resampling to estimate uncertainty
in the differences between forecasting methods. On PVDAQ,
circular seven-day blocks are sampled from the 305 test dates,
with the same dates retained for each paired comparison.
For the external benchmark, a reported model mean first scores each
seed's predictions on the same daylight origin--horizon mask and then
averages those seed errors. The table's mean difference is the
comparator mean minus the STR mean. The paired difference instead
averages each model's seed predictions at every origin and horizon,
scores those mean predictions on the same mask, and subtracts STR
error from comparator error. Both differences are multiplied by 100
for display. In symbols, with $e(\mathbf p)$ denoting the masked,
equal-horizon MAE, these estimands are
\begin{align}
 \Delta_{\rm mean}&=100\left[\frac{1}{S_c}\sum_s e(\mathbf p_{c,s})
 -\frac{1}{S_{\rm STR}}\sum_s e(\mathbf p_{{\rm STR},s})\right],\\
 \Delta_{\rm pair}&=100\left[e(\overline{\mathbf p}_c)
 -e(\overline{\mathbf p}_{\rm STR})\right].
\end{align}
The external paired intervals use 5,000 circular seven-day block
draws. The candidate backbones have three independently selected
seed checkpoints, whereas the three STR adapter seeds share one
frozen TimeMixer checkpoint. Thus the pairing is by forecast origin
and mask, not by matched checkpoint seed. Since absolute error is
nonlinear, the two differences can even have opposite signs.
The backbone-transfer experiment likewise compares three-seed
mean adapter predictions with each matched frozen backbone and
three-seed residual control; positive baseline-minus-STR or
control-minus-STR differences favor STR.

For the weather experiment, paired confidence intervals
are obtained from 4,000 block-resampling draws after
averaging predictions across the three seeds.

To examine performance under different operating conditions,
recent-power volatility is calculated as the mean absolute
first difference over the latest 12 observations. The thresholds
used to define volatility groups are determined from the
training data.

\subsection{Efficiency}

We evaluate the computational overhead introduced by STR in
terms of trainable parameters, training time and inference
latency. Since the backbone remains frozen, we report the
additional parameters and training time required by the
adapter separately from those of the complete forecasting
system.

Inference latency is measured with a batch size of one,
using 30 warm-up runs followed by 200 timed repetitions.
The backbone and its STR-enhanced version are evaluated
in the same session to measure the additional inference
time introduced by the adapter.

Training times are reported according to their original
experimental settings. Latency results for GPU-based neural
models and CPU-based tree models are presented separately,
as their execution times are not directly comparable across
different hardware configurations.

\section{Results}
\subsection{Structured routing versus matched residual correction}
STR has lower dataset-specific scaled MAE than the nominally parameter-matched residual adapter on all four tasks (Table~\ref{tab:mechanism}). Each paired 95\% interval excludes zero. This comparison isolates the implemented trajectory design from the broader benchmark of complete forecasting methods.

\begin{table}[!htbp]
\centering\small
\caption{STR versus the matched residual control. Entries are dataset-specific scaled MAE multiplied by 100; reductions and intervals use the same units. GEFCom uses source-normalized power, Gatton and PVDAQ use nameplate-normalized MAE, and Solar-Energy uses train-standardized macro MAE. Positive reductions favor STR.}
\label{tab:mechanism}
\resizebox{\linewidth}{!}{\begin{tabular}{lrrrr}
\toprule
Dataset & STR error ($\times100$) & Residual ($\times100$) & Reduction ($\times100$) & Paired 95\% CI ($\times100$) \\
\midrule
GEFCom & 2.3137 & 2.5131 & 0.1994 & [0.1226, 0.2797] \\
Gatton & 5.2732 & 5.2946 & 0.0214 & [0.0118, 0.0328] \\
Solar-Energy & 13.4834 & 13.5478 & 0.0644 & [0.0493, 0.0803] \\
PVDAQ & 4.9255 & 4.9504 & 0.0250 & [0.0182, 0.0325] \\
\bottomrule
\end{tabular}}
\end{table}

\subsection{Transfer across frozen neural backbones}
\begin{figure}[tbp]
\centering
\includegraphics[width=\linewidth]{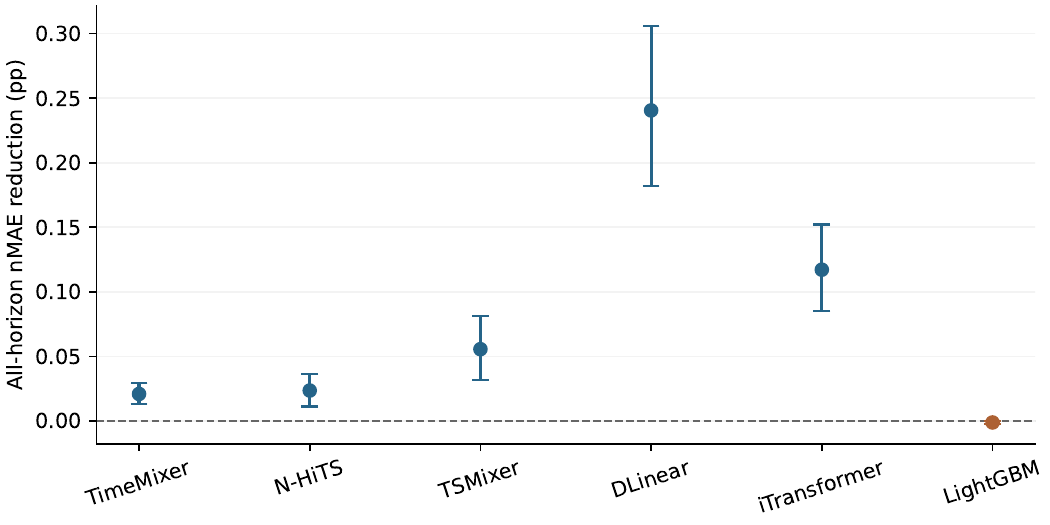}
\caption{Reduction in all-horizon nMAE after adding STR to each frozen PVDAQ backbone, with paired confidence intervals. Positive values indicate lower error with STR; the dashed line marks zero. LightGBM is shown separately from the five neural backbones.}
\label{fig:transfer}
\end{figure}

We next examine whether the same STR design can be used with different frozen neural backbones. On PVDAQ, adding STR reduces all-horizon nMAE for TimeMixer, N-HiTS, TSMixer, DLinear and iTransformer (Table~\ref{tab:backbone-transfer} and Fig.~\ref{fig:transfer}). The corresponding paired 95\% confidence intervals exclude zero both against each original backbone and against its matched residual adapter. Improvements in normalized accuracy range from 0.0201 to 0.2364 percentage points. Each backbone uses an independently trained STR adapter; the learned weights are not transferred between models.

We also evaluate LightGBM to examine whether the same approach benefits a tree-based model. Its observed accuracy change is $-0.0017$ percentage points, with a paired interval of $[-0.0023,+0.0002]$. Because the interval crosses zero, this experiment does not establish an improvement for LightGBM.
\input{supplementary/backbone_transfer_tightened.tex}

\subsection{Horizon-resolved adaptation}
The benefits of STR vary with forecast horizon (Table~\ref{tab:routed-profile}). For TimeMixer, most of the gain occurs at 15 and 30 min, while the estimates from 90 to 120 min are slightly negative and their intervals cross zero. N-HiTS improves from 30 to 120 min, although its 15-min interval includes zero. DLinear improves at all eight routed horizons. TSMixer shows larger gains at the earliest steps but negative differences at 90, 105 and 120 min; for iTransformer, the gain decreases with lead time and its 120-min interval crosses zero. Thus, a reduction in overall error does not imply an improvement at every individual horizon.
\input{supplementary/routed_profile_tightened.tex}
Only the first eight of the sixteen equally weighted PVDAQ horizons are adapted. Because the remaining eight predictions are unchanged, the all-horizon error reduction is exactly half the mean reduction over the routed horizons, apart from display rounding. This is a direct consequence of the evaluation metric and the bypass rule, not an additional empirical finding.

\subsection{Exact long-horizon preservation}
For all evaluated adapters and seeds, predictions from 135 to 240 min are identical to the corresponding frozen backbone outputs. This confirms that the implementation follows Eq.~\eqref{eq:bypass}. The equality is a property of the architecture, not evidence that STR improves third- or fourth-hour accuracy; any errors in the original backbone forecasts are also retained.

\subsection{External benchmark comparison}
Supplementary Table~S.1 compares the original TimeMixer-based STR system with 12 other learning methods on each task, using each task's own error scale. STR ranks third on GEFCom, first on Gatton, third on Solar-Energy and fifth on PVDAQ. LightGBM leads GEFCom and PVDAQ; XGBoost leads Solar-Energy. These ranks place the adapter among complete forecasting methods without selecting the best adapted backbone after the transfer experiment.

On PVDAQ, the three tree ensembles have the lowest all-horizon nMAE: LightGBM 4.5405\%, XGBoost 4.5753\% and CatBoost 4.6068\% (Supplementary Table~S.2). N-HiTS reaches 4.8116\%, followed by the original TimeMixer-based STR instance at 4.9255\%. TSMixer and TimeMixer reach 4.9427\% and 4.9455\%. The five adapted backbones in the transfer study are separate matched comparisons, not candidates from which this benchmark selected an STR result.

Supplementary Table~S.3 adds origin-paired comparisons. Negative comparator-minus-STR values favor the comparator. LightGBM, XGBoost and CatBoost have intervals below zero; TimeMixer, PatchTST, DLinear and TiDE have intervals above zero. TSMixer and iTransformer change sign between the difference of separately scored seed means and the difference scored after averaging seed predictions. Their paired intervals, like those for TimesNet and TimeXer, cross zero. N-HiTS has a lower mean error, but no frozen paired interval was available.

\subsection{Efficiency and recorded training cost}
Supplementary Table~S.2 reports full-system batch-one latency. Among the GPU models, DLinear and TSMixer have the smallest measured medians. In the benchmark timing session, the same frozen TimeMixer checkpoint takes 17.421 ms alone and 18.327 ms with STR, a difference of medians of 0.906 ms (5.2\%). A separate transfer timing session on the same RTX A6000 and checkpoint set gives 17.563 and 19.052 ms, respectively, a 1.49 ms (8.5\%) difference of medians. Both use batch one, 30 warm-ups and 200 randomized timed calls, but their measurements belong to different sessions. Across the five neural backbones in the transfer session, the added full-system median latency is 1.14--1.49 ms. Tree timings use eight CPU threads and are not ranked against GPU timings. The low fitting cost applies to the adapter after a backbone already exists; it does not make complete-system inference free.

Full training-time records and same-session full-system latency comparisons appear in Supplementary Tables~S.4 and S.5. The seven newly trained models follow the same fitting protocol, whereas the remaining timings come from earlier recorded sessions. The representative STR adapter fit takes 25.8 s with the TimeMixer backbone already trained and frozen. This measures the cost of fitting the adapter only, not the training time of a complete STR forecasting system.

\subsection{Boundary and failure analysis}
Recent-power volatility modifies the observed effect (Supplementary Table~S.6). STR reduces error in all three training-defined groups on GEFCom, Gatton and PVDAQ. Solar-Energy's low-volatility group instead has a 0.000407 increase in train-standardized macro MAE; its middle and high groups improve. Thresholds remain training-derived even where many histories have no measured change.

Because confidence intervals were not frozen for the individual volatility groups, Supplementary Table~S.6 reports only point estimates and sample counts. In the consecutive-block analysis, STR improves on the residual control in all 18 evaluated blocks. Its comparison with the original backbone is positive in all but one PVDAQ block, where the difference is only $-2.6\times10^{-7}$. GEFCom includes one test month, and certified calendar dates are unavailable for Solar-Energy; these results therefore cannot establish stability across seasons.

Several results also limit the scope of the findings. The paired interval for LightGBM crosses zero, and TSMixer shows worse performance at some routed horizons. In the exploratory weather experiment, the added weather information does not produce a stable overall benefit for TimeMixer (Supplementary Section~S3). We retain these outcomes under the original split and routing window rather than adjusting the experiment after observing them.

\section{Discussion}
\subsection{Mechanism and backbone transferability}
STR improves on the implemented residual control across four tasks. The control learns $b+d$, whereas STR exposes the backbone forecast, latest-level persistence and local-slope extrapolation before its additive residual. These paths encode different assumptions about how the observed state may develop; their explicit separation may help the router use a transient ramp without forcing the entire correction into one residual. The nominally matched control has inactive routing logits, so this experiment supports the tested design comparison rather than a claim about every possible residual adapter.

Five neural backbones gain in aggregate on the common PVDAQ queue, each with independently fitted STR weights. DLinear gains more than TimeMixer; one possible explanation is that the frozen models leave different amounts of recent-state error for an output adapter to correct. Model capacity itself was not isolated. LightGBM's paired interval crosses zero, so a reliable improvement is not established for that ensemble. Its feature splits might already capture some relevant interactions, but one tree result cannot define a boundary for all tree models. The external ranks describe the original TimeMixer-based instance and do not change these matched transfer estimates.

\subsection{Horizon-dependent behavior}
Most routed-horizon gains occur early, while TSMixer and some other backbones regress at individual later steps within the routing window. Recent level and slope information may lose value as lead time grows, which is consistent with the mixed horizon profile. The exact bypass at 120 min prevents the adapter from changing 135--240 min predictions. It preserves their existing errors as well as their values; it neither improves third- or fourth-hour accuracy nor establishes 120 min as an optimal cutoff for every task.

\subsection{Computational considerations}
The PVDAQ adapter adds 1,204 trainable parameters and has a representative 25.8-s adapter-only fit after backbone training. Full-system inference still runs the backbone. The benchmark session measures 0.906 ms (5.2\%) additional TimeMixer median latency, and the separate transfer session measures 1.49 ms (8.5\%); the latter session spans 1.14--1.49 ms across the five neural backbones. The added fraction is larger for smaller backbones. These measurements describe computational overhead on the recorded hardware; energy consumption was not measured.

\subsection{Weather conditioning}
DLinear W1 gains under temporal pairing, but one of its three adapter seeds regresses. TimeMixer shows no stable aggregate gain, and W2 and W3 give negative or mixed outcomes. Future weather may help when the frozen backbone leaves errors associated with changing conditions, but these data do not establish that mechanism. The 56 origins with conflicting power and radiation trends offer an exploratory clue only: weather enters the additive residual as well as the routing logits, and this small subset cannot identify which component caused the change. Weather is therefore an optional research extension, not a requirement of STR.

\subsection{Limitations and future work}
The transfer study tests five neural backbones and LightGBM at one PVDAQ site. Solar-Energy's low-volatility group regresses, and the four datasets use different error scales. Their public test periods were exposed during development, so the intervals do not substitute for a new independent evaluation. Archived forecasts follow the study's issue-time rules, but historical operational receipt logs are unavailable; valid time and receipt time remain different concepts. A prospectively held-out study across multiple sites and seasons, with receipt records and direct energy measurements where efficiency is claimed, is needed to assess deployment and broader transfer.

\section{Conclusions}
We introduced STR to adapt the short-horizon output of a frozen PV forecasting model. It combines the original forecast with trajectories based on the latest measured power and recent power trend, while preserving all predictions beyond 120 min.

STR reduces error relative to the matched residual adapter on four public PV tasks. On PVDAQ, separately trained STR adapters also improve five different neural backbones, with positive paired confidence intervals for their aggregate results. These experiments establish reuse of the STR design across the tested backbones, not transfer of trained router weights.

No reliable gain is observed for LightGBM, and some volatility groups and individual horizons show worse results. Direct weather conditioning also provides no consistent improvement across the tested backbones. STR is therefore a short-horizon adaptation option for the neural models evaluated here, rather than a replacement forecasting architecture or a method shown to benefit all model families.

\section*{Data availability}
PVDAQ measurements are available through the Open Energy Data Initiative \cite{pvdaq}, and historical GFS forecasts through the NCAR Geoscience Data Exchange \cite{gfs}. GEFCom2014 is described by Hong et al.\ \cite{hong}. The reproducibility package documents the Gatton and Solar-Energy source manifests and preprocessing steps. Derived code and prediction tables will be released after the relevant source licenses and repository metadata have been reviewed.

\section*{Funding}
This research received no external funding.

\section*{Declaration of generative AI and AI-assisted technologies in the manuscript preparation process}
ChatGPT assisted with language revision. The authors reviewed the manuscript and remain responsible for its content.


\bibliographystyle{elsarticle-num}
\renewcommand{\bibfont}{\small}
\bibliography{references}
\end{document}

%% file: supplementary/backbone_transfer_tightened.tex
\begin{table}[tbp]\centering\small
\caption{PVDAQ transfer of the STR formulation. Baselines are matched frozen checkpoints; residual and STR errors are means of three adapter seeds. Positive differences favor STR. Paired intervals use seed-averaged predictions and can differ from differences of separately scored seed means.}
\label{tab:backbone-transfer}
\begin{tabular}{lrrrr}\toprule
Backbone & Base (\%) & Residual (\%) & STR (\%) & Added parameters\\\midrule
TimeMixer & 4.9455 & 4.9504 & 4.9255 & 1,204\\
N-HiTS & 4.8116 & 4.8022 & 4.7888 & 1,204\\
TSMixer & 4.9492 & 4.9387 & 4.8950 & 1,204\\
DLinear & 5.7695 & 5.7381 & 5.5330 & 1,204\\
iTransformer & 4.9934 & 4.9718 & 4.8789 & 1,204\\
\midrule\multicolumn{5}{l}{Non-neural boundary case}\\
LightGBM & 4.5405 & 4.5412 & 4.5421 & 1,204\\\bottomrule\end{tabular}
\par\medskip
\resizebox{\linewidth}{!}{\begin{tabular}{lrrrr}\toprule
Backbone & Base--STR (pp) & Paired 95\% CI & Residual--STR (pp) & Paired 95\% CI\\\midrule
TimeMixer & +0.0201 & [+0.0133,+0.0292] & +0.0250 & [+0.0188,+0.0328]\\
N-HiTS & +0.0229 & [+0.0113,+0.0364] & +0.0134 & [+0.0075,+0.0199]\\
TSMixer & +0.0541 & [+0.0320,+0.0815] & +0.0437 & [+0.0265,+0.0627]\\
DLinear & +0.2364 & [+0.1821,+0.3057] & +0.2050 & [+0.1588,+0.2638]\\
iTransformer & +0.1145 & [+0.0849,+0.1522] & +0.0929 & [+0.0682,+0.1242]\\
\midrule\multicolumn{5}{l}{Non-neural boundary case}\\
LightGBM & -0.0017 & [-0.0023,+0.0002] & -0.0010 & [-0.0012,+0.0000]\\\bottomrule\end{tabular}}\end{table}

%% file: supplementary/routed_profile_tightened.tex
\begin{table}[tbp]\centering\scriptsize
\caption{Frozen routed-horizon nMAE reductions (percentage points) on PVDAQ. Positive values favor STR. Full paired intervals and relative improvements are retained in the supplementary CSV. These mean-prediction profile estimates are distinct from the mean-of-seed metrics in the transfer summary.}
\label{tab:routed-profile}
\resizebox{\linewidth}{!}{\begin{tabular}{lrrrrrrrr}\toprule
Backbone & 15 & 30 & 45 & 60 & 75 & 90 & 105 & 120\\\midrule
TimeMixer & +0.2843 & +0.0472 & +0.0097 & +0.0061 & +0.0021 & -0.0007 & -0.0054 & -0.0073\\
N-HiTS & -0.0082 & +0.0521 & +0.0608 & +0.0528 & +0.0499 & +0.0414 & +0.0498 & +0.0798\\
TSMixer & +0.7846 & +0.2475 & +0.0796 & +0.0010 & -0.0230 & -0.0486 & -0.0671 & -0.0834\\
DLinear & +0.5954 & +0.5048 & +0.5196 & +0.5120 & +0.5182 & +0.4660 & +0.4164 & +0.3155\\
iTransformer & +0.8984 & +0.4369 & +0.2464 & +0.1436 & +0.0806 & +0.0480 & +0.0234 & -0.0021\\
LightGBM & -0.0022 & +0.0013 & +0.0001 & -0.0026 & -0.0007 & -0.0019 & -0.0057 & -0.0057\\\bottomrule\end{tabular}}\end{table}